\documentclass[journal]{ASD}
\IEEEoverridecommandlockouts

\usepackage[english]{babel}
\usepackage[utf8]{inputenc} 
\usepackage{mwe}
\usepackage{caption}
\usepackage{blindtext}
\usepackage{comment}
\usepackage{flushend}
\usepackage{nidanfloat}
\usepackage{amssymb}
\usepackage{subfigure}
\usepackage{booktabs}
\usepackage{chngcntr}

\renewcommand{\thesection}{\arabic{section}} 
\renewcommand{\thefigure}{\arabic{figure}} 
\renewcommand{\thetable}{\arabic{table}}

\usepackage{cite}
\usepackage[cmex10]{amsmath}
\usepackage{multirow}
\usepackage{array}
\usepackage[lofdepth,lotdepth]{subfig}
\usepackage{adjustbox}
\usepackage[labelsep=period]{caption}
\begin{document}


%
\title{Statistical Analysis of Time-Frequency Features Based On Multivariate Synchrosqueezing Transform for Hand Gesture Classification \\ El Hareketi Sınıflandırması için Çok-Değişkenli Senkrosıkıştırma Dönüşümüne Dayalı Zaman-Frekans Özniteliklerinin İstatistiksel Analizi
}

\author{\IEEEauthorblockN{Lutfiye SARIPINAR$^{1}$, Deniz Hande KISA$^{1}$, Mehmet Akif OZDEMIR$^{1}$, Onan GUREN$^{1,*}$}\\
\IEEEauthorblockA{$^{1}$Department of Biomedical Engineering, Izmir Katip Celebi University, Izmir, Turkey\\
ORCIDs: 0000-0001-7137-9271, 0000-0002-5882-0605, 0000-0002-8758-113X, 0000-0001-5808-3767\\
e-mails: lutfiyee35@gmail.com, denizhandekisa@hotmail.com, makif.ozdemir@ikcu.edu.tr, onan.guren@ikcu.edu.tr \\
$^{*}$Corresponding author.}
}


%

\maketitle

\begin{ozet}
Bu çalışmada dört ortak zaman-frekans (TF) momenti olan ve Çok-Değişkenli Senkrosıkıştırma Dönüşümünden (MSST) elde edilen zaman-frekans (TF) matrisinin ortalaması, varyansı, çarpıklığı ve basıklığı, el hareketi tanımada kullanılmak üzere öznitelik olarak önerilmiştir. 10 el hareketi gerçekleştiren 40 katılımcıdan elde edilen yüzey EMG (sEMG) sinyallerini içeren halka açık veri seti kullanıldı. Test edilen hareketler için kullanılan öznitelik değişkenlerinin ayırt edebilme gücü, Kruskal-Wallis (KW) testinden elde edilen $\textbf{\textit{p}}$ değerlerine göre değerlendirildi. TF matrislerine ait ortalama, varyans ve çarpıklığın el hareketlerinin tanınması için aday birer öznitelik kümesi olabileceği sonucuna varılmıştır.


\end{ozet}
\vspace{-1mm}
\begin{IEEEanahtar} 
Çok-değişkenli senkrosıkıştırma dönüşümü, ortak zaman-frekans momenti, elektromiyografi, Kruskal-Wallis testi.

\end{IEEEanahtar}
\begin{abstract}
In this study, the four joint time-frequency (TF) moments; mean, variance, skewness, and kurtosis of TF matrix obtained from Multivariate Synchrosqueezing Transform (MSST) are proposed as features for hand gesture recognition. A publicly available dataset containing surface EMG (sEMG) signals of 40 subjects performing 10 hand gestures, was used. The distinguishing power of the feature variables for the tested gestures was evaluated according to their $\textbf{\textit{p}}$ values obtained from the Kruskal-Wallis (KW) test. It is concluded that the mean, variance, skewness, and kurtosis of TF matrices can be candidate feature sets for the recognition of hand gestures.
\end{abstract}
\vspace{-1mm}



%
\IEEEpeerreviewmaketitle

\IEEEpubidadjcol

\section{Introductıon}
Hand gesture recognition is a procedure for the classification of meaningful hand gestures for human-computer interaction. The surface EMG (sEMG) is an effective and non-invasive method that measures the motor neuron action potentials that are produced by the contraction and relaxation of the muscle during performing the gesture. sEMG is frequently used in hand gesture classification studies as it carries intrinsic information  related to intended gestures \cite{jaramillo2020real}. Hand gesture recognition has a wide real life applications such as smart prostheses, control of rehabilitation devices, exoskeletons, virtual reality, and control of electric-powered wheelchairs \cite{ceolini2020hand,Kisa2020EMGLearning}.

EMG data is processed and analyzed before being used in gesture-based systems, which results in a significant increase in system performance. Time-frequency (TF) analysis (TFA) provides both time and frequency analysis simultaneously. In TFA, the signal is characterized on the T versus F axis that shows the variation of frequency over time and provides non-stationary information of EMG signals \cite{Ozdemir2022Hand,Ozdemir2020EMGLearning}. TFA methods frequently preferred in the processing of biomedical signals can be listed as Short Time Fourier Transform (STFT), Wavelet Transform (WT), Gabor Transform (GT), and Continuous Wavelet Transform (CWT) \cite{kiymik2005comparison}.

Synchrosqueezing Transform (SST) is an effective TFA method that generates a focused and condensed representation of signals in the time-frequency domain (TFD). Multivariate Synchrosqueezing Transform (MSST) is an extension of the SST to analysis of multi-channel multivariate data. The MSST produces dense TF distribution formed from common components obtained from different channels \cite{ahrabian2015synchrosqueezing}.

Feature extraction is one of the basic steps in machine learning based classification studies and aims to represent the signal in  the most informative minimal form. The features obtained from the sEMG data can be extracted from the time domain (TD), frequency domain (FD), or TF domain. The interpreting the TFD as a TF matrix (TFM) and extracting features from the TFM has been proposed as an alternative way for hand gesture classification, recently \cite{rabin2020classification}. Feature selection which follows the feature extraction step aims to select the distinctive features between the different hand gestures. 

Statistical analysis of extracted features is an effective way to select the most distinctive features. The Kruskal-Wallis (KW) test is one of the statistical analysis methods, which is used in hand gesture classification studies \cite{ostertagova2014methodology}. In the KW test, probability values are calculated by using sequencing information to test the distinguishing power of features \cite{fatimah2021hand}.

This study aims to analyze the significance of features extracted from sEMG signals to be used in the classification of hand gestures. The publicly available dataset including sEMG signals of $10$ hand gestures was used. MSST was applied to the sEMG signals to obtain the time-frequency matrices (TFM). The first four order moments of TFM were proposed as a feature  and the statistical significance of each feature for $10$ gestures was analyzed with the KW test. The analyses involve comparing gestures both among themselves and individually.

\section{Related Works}
In the literature, there have been few studies analyzing or classifying hand gesture by using TF matrices from sEMG signals. Li et al. \cite{li2020gesture} worked with $4$-channel sEMG in their study, where they proposed a new method to increase the success of gesture recognition. TFM obtained by applying S-transform (ST) to sEMG signals and they were divided into $16$ sub-matrices along the time and frequency axes. The features were extracted by applying multiscale Singular Value Decomposition (SVD) to all sub-matrices. $9$ different gestures were classified by the Deep Belief Network (DBN) fed with the extracted features.

Fajardo et al. \cite{fajardo2021emg} proposed the combination of handcrafted features from a TFA to recognize ten hand gestures. Huang et al. \cite{huang2019surface} classified $50$ gestures using the TFD features of the spectrograms. There are also studies using TFM-based features using other data types other than EMG. For instance, Lie et al. \cite{li2018effect} used $3$ empirical micro-Doppler features of the TF spectrograms from sparsity-driven TFA to classify dynamic hand gestures.

\section{Materials and Methods}
\subsection{Dataset}\label{Data Acquisition} 
\begin{figure}[htbp]
\centering
\includegraphics[width=1.0\columnwidth]{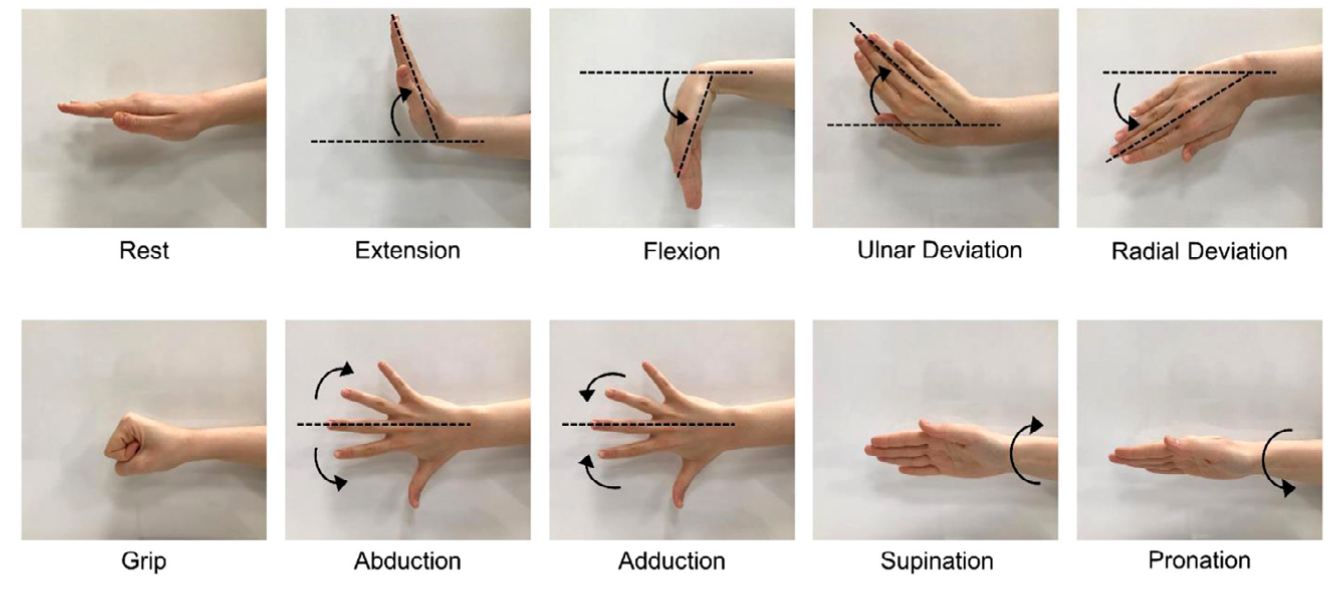}
\caption{The tested ten hand gestures 
\cite{ozdemir2022dataset}.}
\label{fig1:gestures}
\vspace{-6mm}
\end{figure}

In this study, we used a publicly available sEMG data-set which is created by recording $6$ s sEMG signals performed $10$ hand gestures by $40$ healthy participants ($20$ males, $20$ females) \cite{ozdemir2022dataset}. The studied gestures are demonstrated in Fig. \ref{fig1:gestures}. The dataset recorded with the $4$-channel sEMG method was created by collecting the data of each individual in $5$ repetitions in total. The dataset with a sampling rate of $2$ kHz was recorded from the dominant forearm. 

\subsection{Pre-processing and Segmentation}\label{Pre-processing} 
In order to eliminate the noise originated from the external environment or internal organs, all the signals were filtered with the sixth-order Butterworth band-pass filter ($5$-$500$ Hz) and second-order Notch filter ($50$ Hz) were applied with the BIOPAC software during the experiment \cite{ozdemir2022dataset}. Later, $250$ ms gesture were moments extracted from the $6$ s signals applying overlapped sliding window \cite{lee2021electromyogram}. In addition, as a common approach, $1$ s 'transient-states' at the beginning and end were removed from the $6$ s segment, and the middle $4$ s was taken as the 'steady-state' to make a sliding window. At the end of this process, there was a total of $76$ sEMG segments for each $4$ s signal.

\subsection{Multivariate Synchrosqueezing Transform}\label{HHT} 
MSST method enables to identify common oscillations of the data that contains multi-channels with multivariate signals by applying many SST operations in succession \cite{yu2018multisynchrosqueezing,ahrabian2015synchrosqueezing}. Firstly, the SST is applied to the each channel of the data  to determine the channel frequency band and instantaneous frequencies (IF). Next, a common frequency band for all channels is determined and the channel's IF and instantaneous amplitude (IA) are added to compute the IFs and IAs of the common frequency band. For detailed explanation the readers refer to \cite{ahrabian2015synchrosqueezing,yu2018multisynchrosqueezing}.

\subsection{Feature Extraction}\label{Feature Extraction} 
In this study, the four statistical parameters of the TF matrix such as mean, variance, skewness, and kurtosis were selected as features to represent the each hand gesture pattern. In a the two-dimensional ($2$D) TF matrix $T(\omega,t)_{MxN}$, $M$ and $N$ represent frequency resolution and the number of samples in EMG, respectively. The selected features were correlated with the first four order moments and were calculated from Equation \ref{eq:moment}:
  \begin{equation} \label{eq:moment}
        <t^n\omega^m> = \int\int t^n\omega^m{P(\omega,t)}dtd\omega
    \end{equation}

where $P(t,f)$ represents probability distribution of TF matrix, $P(t)$ represents the marginal distribution, $n$ and $m$ represent the order of time and the order of frequency, respectively. In the formula of the mean calculated by the Equation \ref{eq:moment}, $<t^n>$ and $<\omega^m> $ represent the first temporal and spectral time-frequency moments \cite{loughlin2000conditional}. It specifies the first-order joint moment (mean) of coefficient $T(t,\omega)$ the time-frequency distribution of $x$. The variance, skewness, and kurtosis are calculated by taking the second, third, and fourth moments, respectively.

\begin{figure*}[htbp!]
    \centering
    \includegraphics[width=1.6\columnwidth]{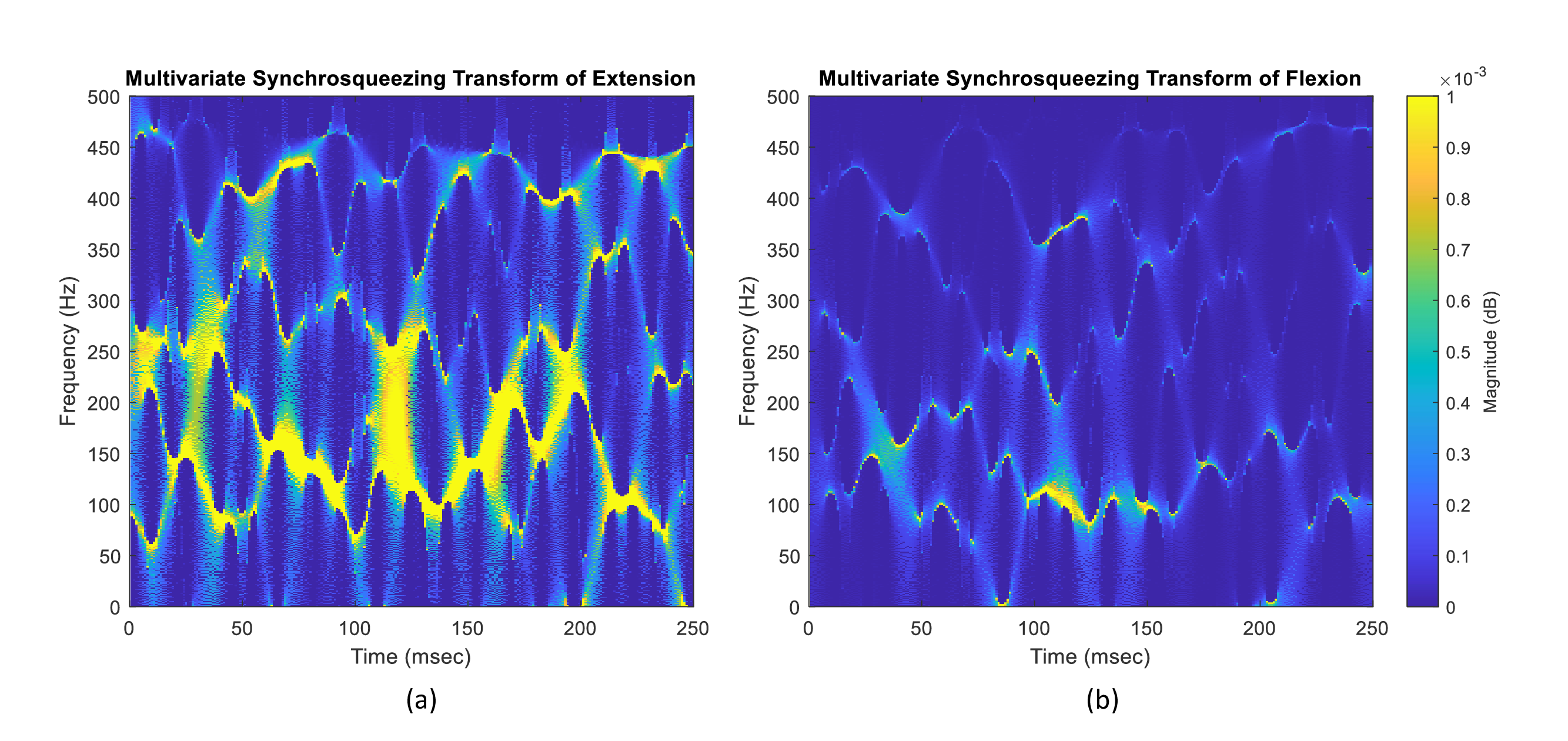}
    \caption{An example of Multivariate Synchrosqueezing Transform of a) extension, and b) flexion movements, respectively.}
    \label{fig2:msst}
\end{figure*}
\vspace{1mm}

\subsection{Statistical Analysis}\label{Statistical}
The KW test is the non-parametric equivalent of the parametric one-way ANOVA test and does not require normal distribution within the groups. KW test examines the significance of the difference between the median values \cite{guo2013privacy} of more than two groups on a certain variable which indicates mean, variance, skewness, and kurtosis in our study. The statistical significance is represented by $p$ values. If the $p$ value of the variable is close to zero, the groups can be separated from each other regarding the variable \cite{fatimah2021hand}. Practically, an appropriate threshold value is chosen and the variables having a $p$ value lower than the threshold are accepted statistically significant. 

\section{Results and Discussion}

In this study, the features extracted from the TF matrix which resulted from MSST of 4-channel sEMG signal were analyzed in terms of statistical significance by using KW rankings. Firstly, $76$ sEMG segments were obtained with a $250$ ms window and $50$ ms increment using the overlapping window approach from $4$ s steady-state signals in the segmentation phase. TF matrices were obtained by applying MSST to $250$ ms four-channel sEMG signals. Thanks to the MSST method, the $4$-channel sEMG signal is represented by a single TF matrix, which provides to preserve the whole information of all channels. In Fig. \ref{fig2:msst}, TFR's of extension and flexion resulted from MSST of $250$ ms segment of the signal are shown, respectively. As shown in Fig. \ref{fig2:msst}, different gestures have different power distributions in their TFR's which leads us to use the distinct properties of the TF matrix as a feature for distinguishing the hand gestures. For this aim,  four features, i.e. mean, variance, skewness, and kurtosis are computed and separated into $10$ groups according to the tested hand gestures. Before statistical analysis, z-score feature normalization was applied to four TF moment features. Statistical analysis was performed with the KW test and the features having $p<0.001$ are assessed as statistically significant \cite{fatimah2021hand}. All processes of the study were carried out in \textsc{Matlab}$^{\circledR}$ environment.

We performed the KW tests on the four features according to inter-subject and intra-subject scenarios. In the inter-subject scenario, the whole data set was examined to determine statistical significance between every feature. In the intra-subject scenario, the data of a different number of the subjects were combined and the statistical analyses were performed separately.   
\begin{figure*}[htbp!]
    \centering
    \includegraphics[width=1.8\columnwidth]{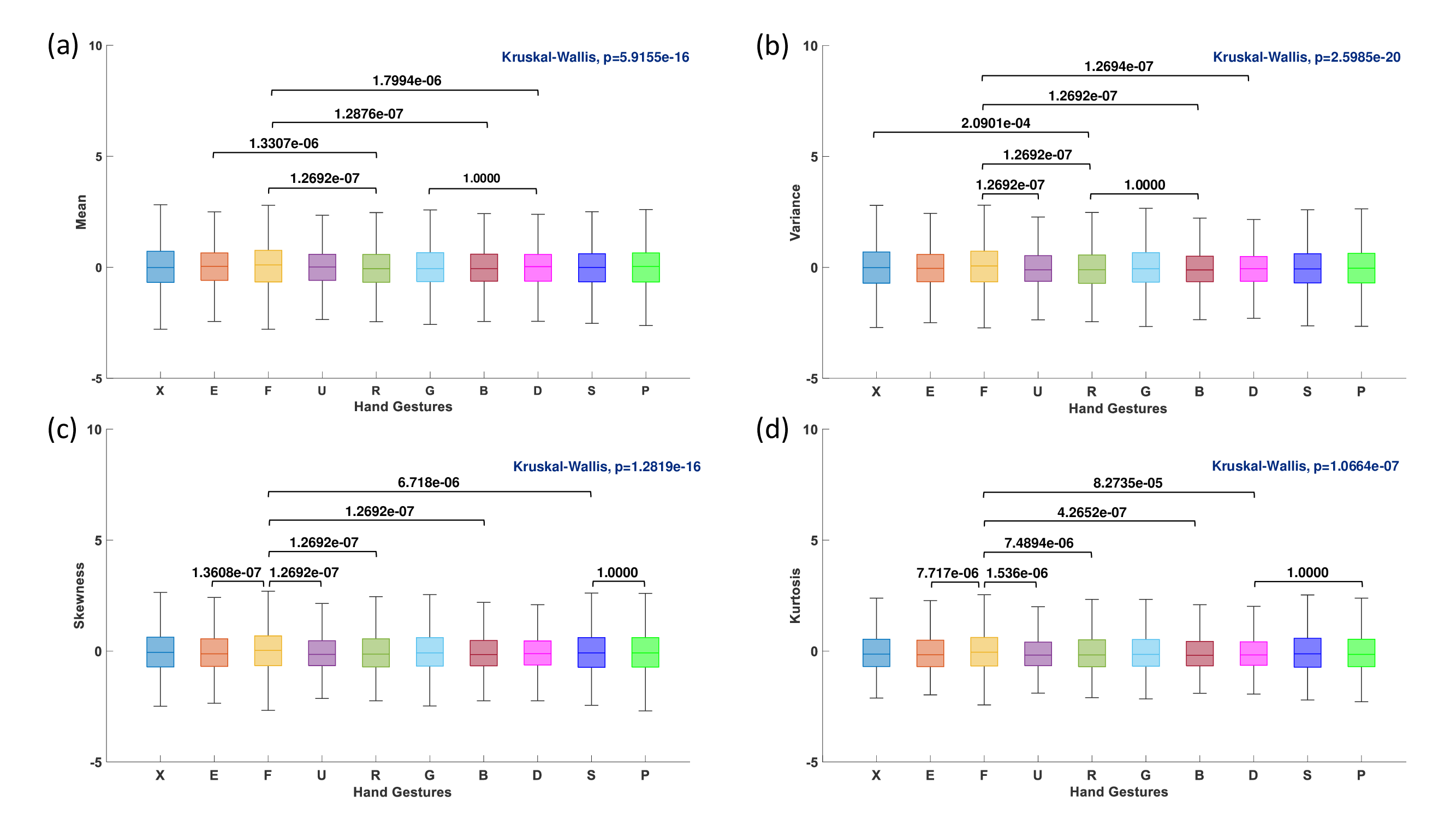}
    \caption{The box plots and the highest and lowest $p$ values of each feature for all subjects a) mean, b) variance, c) skewness, and d) kurtosis, respectively (X: Rest, E: Extension, F: Flexion, U: Ulnar Deviation, R: Radial Deviation, G: Grip, B: fingers abduction, D: fingers adduction, S: supination, and P: pronation).}
\end{figure*}
\label{fig3:boxplots}
\vspace{1mm}

\begin{figure*}[htbp!]
    \centering
    \includegraphics[width=1.8\columnwidth]{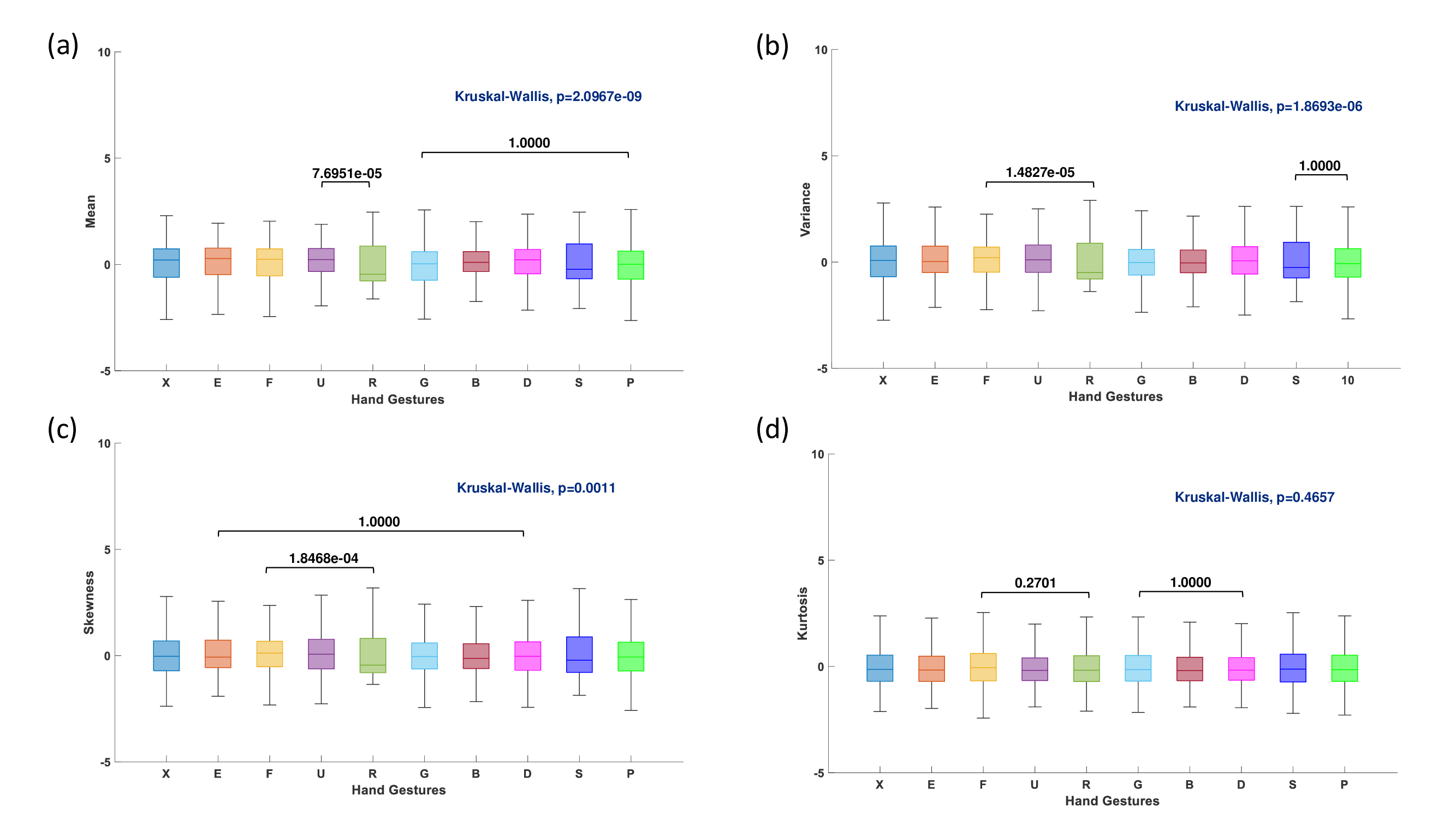}
    \caption{The box plots and the highest and lowest $p$ values of each feature for 5 subjects: a) mean, b) variance, c) skewness, and d) kurtosis, respectively (X: Rest, E: Extension, F: Flexion, U: Ulnar Deviation, R: Radial Deviation, G: Grip, B: fingers abduction, D: fingers adduction, S: supination, and P: pronation).}
\end{figure*}
\label{fig4:boxplot5}
\vspace{1mm}

The statistical analysis results of the inter-subject scenario are shown in Fig. \ref{fig3:boxplots}. The $p$ values for mean, variance, skewness, and kurtosis were obtained as $5.9155e$-$16$, $2.5985e$-$20$, $1.2819e$-$16$ and $1.0664e$-$07$, respectively. The variance with the smallest $p$ value is found as the most significant feature for the classification of tested hand gestures. Moreover, $45$ binary combinations of $10$ gestures for each feature were tested and some of the lowest and the highest $p$ values were also shown in Fig. \ref{fig3:boxplots}.  The ranges (minimums and maximums) of $p$ values are obtained as [$1.2692e$-$07$, $1$] for mean, variance, and skewness, and [$4.2652e$-$07$, $1$] for kurtosis. For the mean, the best significance value  was obtained as $1.2692e$-$07$ between flexion and radial deviation, and there was no significant difference between grip and adduction gestures. For the variance and skewness, the best significance value  was obtained as $1.2692e$-$07$ between flexion and ulnar deviation, flexion and radial deviation. 
There were no significant differences between the radial deviation and abduction gestures in terms of variance as well as the supination and pronation gestures in terms of skewness. The flexion and abduction gestures differentiate in terms of kurtosis $p=4.2652e$-$07$ significantly, whereas the adduction and pronation gestures do not differentiate. 

The flexion gesture has the most different signal pattern in the tested gestures \cite{Ozdemir2022Hand}. TFR matrix of flexion represents this property by generating the most significantly different features. The mean, variance, skewness, and kurtosis of flexion have the lowest $p$ values in all analysis combinations.

Intra-subject scenario was carried out to evaluate the effect of the number of data and subjects on the $p$ value. The statistical significance of the four features for each of the $40$ subjects was investigated separately and it is observed that $p$ value of all features is bigger than $0.001$ for ten hand gestures groups. The average $p$ values of the mean, variance, skewness, and kurtosis were found $0.7231$, $0.7841$, $0.9034$ and $0.9820$, respectively. The same analyses were performed with combination of $2$-subject, $3$-subject and $4$-subject data and the same statistical inference was obtained. The statistical significance of mean, variance, and skewness begins when the data of $5$ subjects were analyzed and their $p$ values decreased to $2.0967e$-$09$, $1.8693e$-$06$, $0.0001$, respectively, as seen in Fig. \ref{fig4:boxplot5}. The kurtosis needs the data of $30$ subjects to be significant statistically.

It is well known that in the machine learning era, feature extraction and feature selection steps affect the performance of the method used for classification \cite{geethanjali2016mechatronics}. Statistical analysis of the extracted features provides to ensure selecting the most distinctive features with the minimum number. Our results propose that each of feature extracted from TFM is an informative candidate for the following steps. in the case of  sufficient data is provided. We claim that MSST plays a critical role in the success of the method by carrying information of all channels into TFM, which needs to be tested for the other TFR methods.  

\section{Conclusion}
In the study conducted to investigate the statistical significance of extracted joint moment features of the TF matrix obtained with MSST method. The statistical analyses were performed with the KW test. The features mean, variance, and skewness were found meaningful for the discrimination of hand sEMG signals of hand gestures. The kurtosis was assessed  less significant than the others. 
The next step of the study is the testing classification performance of the extracted features in classification of hand gesture. In addition, new different features will be extracted and statistical analysis will be performed.

\section*{ACKNOWLEDGEMENT}
This work was supported by the IKCU Scientific Research Projects Coordination Unit [grant numbers $2021$-ÖDL-MÜMF-$0004$, $2022$-GAP-MÜMF-$0001$].




\bibliographystyle{IEEEtran}

\bibliography{asd_makale.bib}

\begin{thebibliography}{10}
\providecommand{\url}[1]{#1}
\csname url@samestyle\endcsname
\providecommand{\newblock}{\relax}
\providecommand{\bibinfo}[2]{#2}
\providecommand{\BIBentrySTDinterwordspacing}{\spaceskip=0pt\relax}
\providecommand{\BIBentryALTinterwordstretchfactor}{4}
\providecommand{\BIBentryALTinterwordspacing}{\spaceskip=\fontdimen2\font plus
\BIBentryALTinterwordstretchfactor\fontdimen3\font minus
  \fontdimen4\font\relax}
\providecommand{\BIBforeignlanguage}[2]{{%
\expandafter\ifx\csname l@#1\endcsname\relax
\typeout{** WARNING: IEEEtran.bst: No hyphenation pattern has been}%
\typeout{** loaded for the language `#1'. Using the pattern for}%
\typeout{** the default language instead.}%
\else
\language=\csname l@#1\endcsname
\fi
#2}}
\providecommand{\BIBdecl}{\relax}
\BIBdecl

\bibitem{jaramillo2020real}
A.~Jaramillo-Y{\'a}nez, M.~E. Benalc{\'a}zar, and E.~Mena-Maldonado,
  ``Real-time hand gesture recognition using surface electromyography and
  machine learning: A systematic literature review,'' \emph{Sensors}, vol.~20,
  no.~9, p. 2467, 2020.

\bibitem{ceolini2020hand}
E.~Ceolini, C.~Frenkel, S.~B. Shrestha, G.~Taverni, L.~Khacef, M.~Payvand, and
  E.~Donati, ``Hand-gesture recognition based on emg and event-based camera
  sensor fusion: A benchmark in neuromorphic computing,'' \emph{Frontiers in
  neuroscience}, vol.~14, p. 637, 2020.

\bibitem{Kisa2020EMGLearning}
D.~H. Kisa, M.~A. Ozdemir, O.~Guren, and A.~Akan, ``{EMG based Hand Gesture
  Classification using Empirical Mode Decomposition Time-Series and Deep
  Learning},'' in \emph{Medical Technologies National Congress
  (TIPTEKNO)}.\hskip 1em plus 0.5em minus 0.4em\relax Antalya, Turkey, 2020,
  pp. 1--4.

\bibitem{Ozdemir2022Hand}
M.~A. Ozdemir, D.~H. Kisa, O.~Guren, and A.~Akan, ``Hand gesture classification
  using time–frequency images and transfer learning based on cnn,''
  \emph{Biomedical Signal Processing and Control}, vol.~77, p. 103787, 2022.

\bibitem{Ozdemir2020EMGLearning}
M.~A. Ozdemir, D.~H. Kisa, O.~Guren, A.~Akan, and A.~Onan, ``{EMG based Hand
  Gesture Recognition using Deep Learning},'' in \emph{Medical Technologies
  National Congress (TIPTEKNO)}.\hskip 1em plus 0.5em minus 0.4em\relax
  Antalya, Turkey, 2020, pp. 1--4.

\bibitem{kiymik2005comparison}
M.~K. K{\i}ym{\i}k, {\.I}.~G{\"u}ler, A.~Dizib{\"u}y{\"u}k, and M.~Ak{\i}n,
  ``Comparison of stft and wavelet transform methods in determining epileptic
  seizure activity in eeg signals for real-time application,'' \emph{Computers
  in biology and medicine}, vol.~35, no.~7, pp. 603--616, 2005.

\bibitem{ahrabian2015synchrosqueezing}
A.~Ahrabian, D.~Looney, L.~Stankovi{\'c}, and D.~P. Mandic,
  ``Synchrosqueezing-based time-frequency analysis of multivariate data,''
  \emph{Signal Processing}, vol. 106, pp. 331--341, 2015.

\bibitem{rabin2020classification}
N.~Rabin, M.~Kahlon, S.~Malayev, and A.~Ratnovsky, ``Classification of human
  hand movements based on emg signals using nonlinear dimensionality reduction
  and data fusion techniques,'' \emph{Expert Systems with Applications}, vol.
  149, p. 113281, 2020.

\bibitem{ostertagova2014methodology}
E.~Ostertagova, O.~Ostertag, and J.~Kov{\'a}{\v{c}}, ``Methodology and
  application of the kruskal-wallis test,'' in \emph{Applied Mechanics and
  Materials}, vol. 611.\hskip 1em plus 0.5em minus 0.4em\relax Trans Tech Publ,
  2014, pp. 115--120.

\bibitem{fatimah2021hand}
B.~Fatimah, P.~Singh, A.~Singhal, and R.~B. Pachori, ``Hand movement
  recognition from semg signals using fourier decomposition method,''
  \emph{Biocybernetics and Biomedical Engineering}, vol.~41, no.~2, pp.
  690--703, 2021.

\bibitem{li2020gesture}
W.~Li, Z.~Luo, Y.~Jin, and X.~Xi, ``Gesture recognition based on multiscale
  singular value entropy and deep belief network,'' \emph{Sensors}, vol.~21,
  no.~1, p. 119, 2020.

\bibitem{fajardo2021emg}
J.~M. Fajardo, O.~Gomez, and F.~Prieto, ``Emg hand gesture classification using
  handcrafted and deep features,'' \emph{Biomedical Signal Processing and
  Control}, vol.~63, p. 102210, 2021.

\bibitem{huang2019surface}
D.~Huang and B.~Chen, ``Surface emg decoding for hand gestures based on
  spectrogram and cnn-lstm,'' in \emph{2019 2nd China Symposium on Cognitive
  Computing and Hybrid Intelligence (CCHI)}.\hskip 1em plus 0.5em minus
  0.4em\relax IEEE, 2019, pp. 123--126.

\bibitem{li2018effect}
G.~Li, S.~Zhang, F.~Fioranelli, and H.~Griffiths, ``Effect of sparsity-aware
  time--frequency analysis on dynamic hand gesture classification with radar
  micro-doppler signatures,'' \emph{IET Radar, Sonar \& Navigation}, vol.~12,
  no.~8, pp. 815--820, 2018.

\bibitem{ozdemir2022dataset}
M.~A. Ozdemir, D.~H. Kisa, O.~Guren, and A.~Akan, ``Dataset for multi-channel
  surface electromyography (semg) signals of hand gestures,'' \emph{Data in
  brief}, vol.~41, p. 107921, 2022.

\bibitem{lee2021electromyogram}
K.~H. Lee, J.~Y. Min, and S.~Byun, ``Electromyogram-based classification of
  hand and finger gestures using artificial neural networks,'' \emph{Sensors},
  vol.~22, no.~1, p. 225, 2021.

\bibitem{yu2018multisynchrosqueezing}
G.~Yu, Z.~Wang, and P.~Zhao, ``Multisynchrosqueezing transform,'' \emph{IEEE
  Transactions on Industrial Electronics}, vol.~66, no.~7, pp. 5441--5455,
  2018.

\bibitem{loughlin2000conditional}
P.~Loughlin, F.~Cakrak, and L.~Cohen, ``Conditional moments analysis of
  transients with application to helicopter fault data,'' \emph{Mechanical
  Systems and Signal Processing}, vol.~14, no.~4, pp. 511--522, 2000.

\bibitem{guo2013privacy}
S.~Guo, S.~Zhong, and A.~Zhang, ``Privacy-preserving kruskal--wallis test,''
  \emph{Computer methods and programs in biomedicine}, vol. 112, no.~1, pp.
  135--145, 2013.

\bibitem{geethanjali2016mechatronics}
P.~Geethanjali, ``A mechatronics platform to study prosthetic hand control
  using emg signals,'' \emph{Australasian physical \& engineering sciences in
  medicine}, vol.~39, no.~3, pp. 765--771, 2016.

\end{thebibliography}
%



\end{document}